\pdfoutput=1

\documentclass[11pt]{article}

\usepackage{ACL2023}

\usepackage{times}
\usepackage{latexsym}

\usepackage[T1]{fontenc}

\usepackage[utf8]{inputenc}

\usepackage{microtype}

\usepackage{inconsolata}

\usepackage{tikz}

\usepackage{array}
\usepackage{makecell}

\usepackage{amsmath}
\usepackage{amsfonts}
\usepackage{amssymb}
\usepackage{graphicx}
\usepackage{geometry}
\usepackage{natbib}
\usepackage{hyperref}
\usepackage{booktabs}
\usepackage{algorithm}
\usepackage{algorithmic}
\usepackage{multirow}
\usepackage{adjustbox}  
\usepackage[most]{tcolorbox}
\tcbuselibrary{minted,breakable}

\usepackage{pifont, xcolor}
\newcommand{\cmark}{\textcolor{green}{\ding{51}}}
\newcommand{\xmark}{\textcolor{red}{\ding{55}}}

\usepackage{tcolorbox}
\tcbuselibrary{skins} 
\definecolor{todo-yellow}{RGB}{255,253,160}
\definecolor{todo-border}{RGB}{200,200,200} 
\definecolor{todo-text}{RGB}{0,0,150} 

\title{IDP Accelerator: Agentic Document Intelligence from Extraction to Compliance Validation}

\author{\textbf{Md Mofijul Islam, Md Sirajus Salekin, Joe King, Priyashree Roy, Vamsi Thilak Gudi}\\ \textbf{Spencer Romo, Akhil Nooney, David Kaleko, Boyi Xie, Bob Strahan, Diego A. Socolinsky} \\ \textbf{Amazon Web Services}}

\begin{document}

\maketitle

\begin{abstract} 

Understanding and extracting structured insights from unstructured documents remains a foundational challenge in industrial NLP. While Large Language Models (LLMs) enable zero-shot extraction, traditional pipelines often fail to handle multi-document packets, complex reasoning, and strict compliance requirements. We present IDP (Intelligent Document Processing) Accelerator, a framework enabling agentic AI for end-to-end document intelligence with four key components: (1) DocSplit, a novel benchmark dataset\footnote{https://huggingface.co/datasets/amazon/doc\_split} and multimodal classifier using BIO tagging to segment complex document packets; (2) configurable Extraction Module leveraging multimodal LLMs to transform unstructured content into structured data; (3) Agentic Analytics Module, compliant with the Model Context Protocol (MCP) providing data access through secure, sandboxed code execution; and (4) Rule Validation Module replacing deterministic engines with LLM-driven logic for complex compliance checks. The interactive demonstration enables users to upload document packets, visualize classification results, and explore extracted data through an intuitive web interface. We demonstrate effectiveness across industries, highlighting a production deployment at a leading healthcare provider achieving 98\% classification accuracy, 80\% reduced processing latency, and 77\% lower operational costs over legacy baselines. IDP Accelerator is open-sourced\footnote{https://github.com/aws-samples/sample-genai-idp \label{fn:idp_accelerator}} with a live demonstration\footnote{https://github.com/aws-samples/sample-genai-idp/blob/main/docs/demo-videos.md \label{fn:demo}} available to the community. 
\end{abstract}


\section{Introduction}

Unstructured data, including forms, emails, and scanned images, constitutes an estimated $80\%$ to $90\%$ of the world's global data assets~\cite{harbert2021tapping}. This vast repository of information holds immense strategic value, yet organizations struggle to extract actionable insights from it. Traditional document processing approaches have proven inadequate: manual processing remains slow and costly, rule-based automation fails when data deviates from predefined templates, and Optical Character Recognition (OCR) extracts raw text without understanding meaning, structure, or intent~\cite{pingili2025IDP-finance}. These limitations result in operational bottlenecks such as slow loan approvals, delayed claims processing, compliance gaps, and increased exposure to fraud. The banking and finance industry alone processes thousands of documents daily, where a single loan application can span hundreds of pages requiring manual validation at multiple stages~\cite{jansen2023rise}.

With the emergence and evolution of Large Language Models (LLMs) starting in 2020, document intelligence has become one of the fastest-growing and most active research areas~\cite{ke2025llm-document-survey}. Interest surged dramatically from mid-2023, driven by foundational breakthroughs such as the extension of LLM context windows enabling practical long-document understanding. Commercial platforms like Azure Document Intelligence and Amazon Textract with Bedrock now integrate LLMs into document workflows at scale, facilitating intelligent question answering, summarization, and structured information extraction. Domain-adapted models like DocLLM \cite{wang2024docllm}, TableLLM \cite{zhang2025TableLLM}, and InstructDoc \cite{tanaka2024InstructDoc} have emerged for specialized tasks including table reasoning and layout comprehension, while practical pipelines such as RagFlow \cite{ragflow2025github} and MinerU \cite{bin-wang-2024-mineru} have operationalized LLMs for document chunking, retrieval, and question-answering over multimodal documents ~\cite{ke2025llm-document-survey}. Industry adoption has accelerated accordingly. IDP (Intelligent Document Processing)  adoption in financial services grew from $18\%$ in 2020 to $45\%$ in 2023 and a projected $72\%$ by 2025, with similar trajectories across insurance, healthcare, and government sectors~\cite{kapula2025intelligent}. Organizations deploying IDP report processing time reductions exceeding $65\%$, data entry error reductions of up to $90\%$, and $70\%$ less manual review effort enabled by NLP and human-in-the-loop designs \cite{kapula2025intelligent}.

Despite these advances, bridging the gap from prototype to production remains challenging, as systems effective on small document sets often fail at enterprise scale due to inadequate error handling, cost inefficiencies, and unmet security requirements. We introduce the IDP Accelerator~\cite{Strahan2025awsblog}, an open-source framework\footref{fn:idp_accelerator} that enables Agentic AI for production-grade document intelligence, providing the research community with a robust platform for agentic document research and reproducible experimentation. Rather than a one-size-fits-all solution, the system provides a modular, customizable platform with four key capabilities:
1. \textbf{DocSplit}: Segments multi-document packets into constituent documents using BIO tagging~\cite{ramshaw-marcus-1995-text}.
2. \textbf{Entity Extraction Module}: Leverages multimodal LLMs to extract structured fields via customizable attribute definitions.
3. \textbf{Agentic Analytics Module}: Enables natural language querying over processed documents via RAG and MCP integration.
4. \textbf{Rule Validation Module}: Applies LLM-driven logic to complex compliance checks beyond deterministic rule engines.
Additional capabilities are detailed in the Appendix. We demonstrate efficacy through real-world deployments across healthcare, finance, marketing, and technology sectors.

\section{Real-world Impact}
The IDP Accelerator is deployed at scale across healthcare, marketing, finance, and legal industries. A leading genetic testing and precision medicine provider previously processed thousands of medical documents daily using Amazon Comprehend, achieving $94\%$ classification accuracy at high cost and latency. With the IDP Accelerator and Amazon Nova models, they achieved $98\%$ accuracy while reducing costs by $77\%$ and latency by $80\%$, saving $300$ hours monthly across $9,000$ prior authorizations and projecting $\$132K$ in annual savings~\cite{Roy2025amazonblog}. The healthcare success story reflects a broader pattern of impact across diverse sectors. In marketing intelligence, a competitive intelligence firm faced the challenge of processing about $45,000$ campaigns daily while maintaining a searchable archive of $45$ million campaigns spanning $15$ years. Using our IDP Accelerator, they achieved $85\%$ classification and extraction accuracy, removed critical operational bottlenecks and reached production deployment in just $8$ weeks~\cite{Strahan2025awsblog}. For enterprise document management, a global technology services company implemented the system to serve their healthcare clients, processing over $10,000$ documents monthly with potential to scale to $70,000$. Their deployment yields savings of over $1,900$ person-hours annually while automating grievance and appeals classification, demonstrating how the system's modular architecture adapts to specialized industry requirements~\cite{Strahan2025awsblog}. At continental scale, North America's largest community management company used the IDP Accelerator 
to organize $48$ million documents across $26$ terabytes spanning $300+$ branch offices. The system now automatically classifies homeowner association documents including bylaws, contracts and meeting minutes, achieving $95\%$ accuracy across nine document types while reducing processing time from hours to seconds~\cite{David2025Associablog}.

These deployments demonstrate the IDP Accelerator's versatility and production readiness across varying scales and industry requirements. Beyond classification and extraction, the Analytics Agent further extends this impact by enabling non-technical users to query processed documents through natural language. From healthcare compliance to marketing analytics to community management, the IDP Accelerator provides a foundation for organizations to unlock the strategic value embedded in their document workflows.

\begin{figure*}[!ht]
    \centering
    \includegraphics[width=0.8\linewidth]{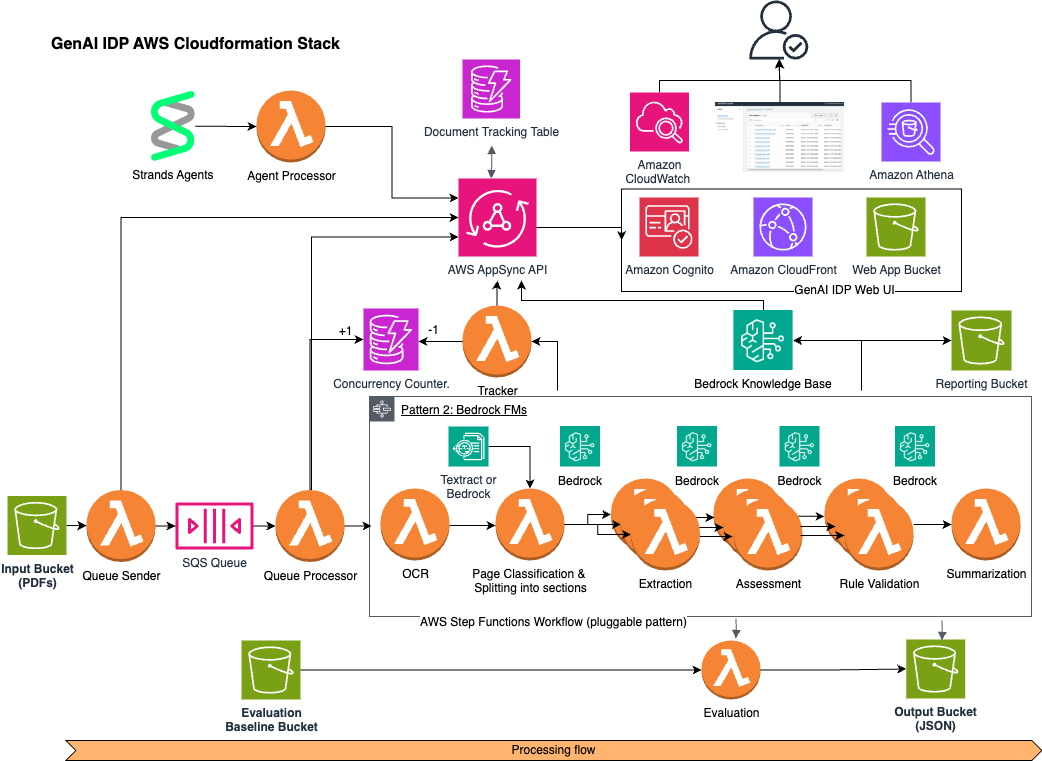}
    \caption{IDP Accelerator system architecture showing the serverless orchestration layer, modular processing patterns, and human-in-the-loop integration points.}
    \label{fig:architecture}
\end{figure*}


\section{Related Systems}

Large language models have been widely used in document processing~\cite{ke2025llm-document-survey, wang2025docai-llm-survey}. Proprietary software systems (e.g., Azure Document Intelligence~\cite{azure-di}, Google Document AI~\cite{google-docai}) provide pre-built solutions with limited information about implementation details and offer limited customization. IDP Accelerator is an open-source solution with a modular architecture that enables researchers and practitioners to extend, customize, and reproduce the system for their specific requirements. 

Agentic AI improves the agility of workflow automation and introduces versatile toolkits to execute a range of IDP tasks ~\cite{vignesh2025idpflow, bhuvaneswaran2025agenticinvoiceprocessing} while enhancing document understanding in long context~\cite{sun-etal-2025-docagent}. 
Building on this paradigm, IDP Accelerator dynamically classifies, splits, extracts, and validates document content, combining an agentic analytics module with LLM-driven components for classification, extraction, and compliance validation, reducing reliance on pre-defined templates or instructions.
It also provides enterprise-grade infrastructure with high scalability, allowing users to focus on business logic.

Scientific experimentation toolkits such as Hugging Face~\cite{wolf-etal-2020-transformers}, spaCy~\cite{honnibal2020spacy}, and scikit-learn~\cite{pedregosa2011scikit} enables research community to experiment with machine learning and LLMs. IDP Accelerator advances this mission by providing a Command Line Interface (CLI) and Test Studio with built-in human-in-the-loop capabilities, enabling researchers to conduct large-scale IDP experiments that accelerate research and experimentation in document processing.



\section{System Architecture}

IDP Accelerator is built on a cloud-native, serverless architecture designed for scalability, cost efficiency, and operational simplicity. Figure~\ref{fig:architecture} illustrates the high-level system architecture.

\textbf{Serverless Orchestration}: The system leverages AWS Step Functions as the central orchestration engine, coordinating document processing workflows as state machines. Individual processing tasks execute as AWS Lambda functions, providing automatic scaling and pay-per-invocation pricing. Asynchronous communication between components is mediated through Amazon SQS queues, enabling decoupled processing stages that can independently scale based on workload. Processing state and extracted data persist in Amazon DynamoDB, ensuring durability and enabling workflow resumption after failures. Amazon CloudWatch provides comprehensive monitoring with detailed metrics and logs for operational visibility.

\textbf{Modular Processing Patterns}: IDP Accelerator implements a pattern-based architecture where document processing workflows are composed from reusable, pre-built processing patterns. The system provides a few core patterns that address different document processing requirements. For example, the Bedrock Data Automation (BDA) pattern leverages Amazon BDA for end-to-end packet and media processing. BDA provides native multimodal understanding, enabling direct processing of complex document formats without intermediate OCR steps. 
On the other hand, the OCR + Bedrock pattern combines OCR with Amazon Bedrock's large language models for extraction. OCR handles text extraction and structural analysis (tables, forms, key-value pairs), while LLMs perform semantic extraction and field mapping. This pattern excels for text-heavy documents requiring precise character-level extraction.
Each pattern can be independently configured and combined to address diverse requirements. Organizations can select existing patterns or extend the system with custom patterns implemented as Lambda functions.

\textbf{Human in the Loop (HITL)}: The architecture incorporates a built-in human review system integrated directly into the Web UI. When document processing confidence falls below configurable thresholds with human review enabled, the system routes documents to the review portal. The system supports role-based access control with Admin and Reviewer personas, enabling organizations to leverage internal domain experts for validation tasks. Review outcomes feed back into the processing pipeline, both completing the immediate workflow and contributing to continuous improvement through correction history and quality metric updates.

\textbf{Web Interface and API Layer}: The solution includes a web interface delivered through CloudFront. Amazon Cognito provides secure authentication with role-based access control, while an AppSync GraphQL API connects the UI to backend services with real-time data subscriptions. The interface supports document upload, status tracking, results visualization, and configuration management. Lambda resolvers handle complex operations including document status lookups and evaluation result retrieval.

\textbf{Monitoring and Observability}: The solution creates integrated CloudWatch dashboards and alerts combining metrics from both core infrastructure and pattern-specific resources. Dashboards track queue performance, workflow statistics, error rates, and resource utilization. Pattern-specific metrics include OCR performance, classification accuracy, and extraction statistics. Centralized logging with correlation IDs enables tracing individual document processing steps.

\textbf{Resilience and Security}: The architecture implements comprehensive error handling with retry strategies in Step Functions state machines. Dead letter queues capture failed messages for manual intervention, while DynamoDB state tracking enables workflow resumption after transient failures. Concurrency management prevents overwhelming downstream services while maximizing throughput. 
Security follows defense principles with S3 encryption using customer-managed KMS keys, least-privilege IAM roles for all Lambda functions, and optional VPC endpoints for private connectivity. Cognito enforces strong password policies with optional multi-factor authentication. All API calls are logged to CloudTrail for audit purposes, and the architecture supports deployment in isolated VPCs for highly regulated environments.

\section{Key System Capabilities}



The IDP Accelerator provides a comprehensive suite of capabilities for enterprise document processing. We highlight the core features below, with emphasis on the NLP and multimodal understanding components that enable robust document intelligence.\footnote{Detailed documentation is available at \url{https://github.com/aws-samples/sample-genai-idp/tree/main/docs}}


\textbf{Multimodal Document Understanding}: IDP Accelerator employs multimodal language models to jointly reason over textual content and visual layout features. The system processes documents as images through Amazon Bedrock-hosted or open-source models to capture spatial relationships, formatting cues, and visual elements that convey semantic information beyond raw text. This multimodal representation enables accurate processing of forms, tables, and multi-column layouts where traditional OCR pipelines lose structural context. For document classification, the system performs page-level analysis to identify document types and boundaries within multi-document packets, enabling automatic segmentation of composite files containing heterogeneous documents. Classification schemas support both predefined taxonomies and custom natural language definitions, allowing domain adaptation without model retraining. The extraction module generates structured JSON output conforming to user-defined schemas, with each field localized through bounding box coordinates, and supports few-shot learning for rapid customization through example-based prompting without fine-tuning. To rigorously evaluate the document splitting capability, we developed \textit{DocSplit}, a comprehensive benchmark dataset alongside novel evaluation metrics for document packet recognition and splitting tasks \cite{islam2026docsplit}.

\textbf{Confidence Estimation}: 
The system employs a multi-layered confidence scoring mechanism across three processing stages. At the OCR layer, AWS Textract generates per-line confidence scores (0–100\%) based on document resolution, font characteristics, and handwritten versus printed content. After the extraction phase, an LLM analyzes each extracted attribute against both OCR-derived text and document imagery, producing granular confidence scores (0.0–1.0) with justifications and spatial localization via bounding box coordinates. When attribute-level confidence falls below a configurable threshold (default: 0.8), the system triggers alerts that may initiate a human-in-the-loop review workflow, ensuring quality assurance for low-confidence extractions.

\textbf{Automated Evaluation Framework}: 
The system provides an evaluation pipeline for document splitting and field extraction, powered by \textit{Stickler}\footnote{https://github.com/awslabs/stickler}, our open-source library for structured object comparison. For extraction, \textit{Stickler} provides extensible attribute-level comparators supporting exact and fuzzy string matching, threshold-based numeric comparison, and bounding box IoU evaluation for spatial elements. We designed \textit{Stickler} with field-level weighting for prioritizing business-critical attributes and Hungarian algorithm-based list matching for accurate comparison of table rows and repeated structures. Results aggregate across document collections as both micro-averaged (field-level) and macro-averaged (document-level) statistics, enabling systematic comparison of model configurations and prompting strategies. For document splitting, the framework evaluates page-level classification accuracy, ordered splitting accuracy with exact page sequence preservation, and unordered splitting accuracy based on section grouping correctness.


\textbf{Agentic Analytics}: Beyond extraction, IDP Accelerator provides agentic analytics capabilities that enable natural language querying over processed document repositories. Users can pose questions about their document collections through a Document Knowledge Base Query interface, which leverages retrieval-augmented generation (RAG) to synthesize answers from relevant extracted content. The RAG pipeline indexes extracted fields and document text, enabling semantic search across document collections followed by LLM-powered answer synthesis. The system integrates with the Model Context Protocol (MCP), enabling external applications such as Amazon Quick Suite to access IDP data and analytics through AWS Bedrock AgentCore Gateway. This extensibility allows organizations to incorporate document intelligence into broader enterprise workflows, business intelligence dashboards, and decision-support systems.

\begin{table*}[!t]
\centering
\small
\caption{Model performance comparison on the RealKIE-FCC-Verified benchmark across two input modalities. Claude Sonnet 4.5 achieves the highest extraction score with OCR+Image, while OCR-based input consistently outperforms image-only across all models, and open-source models (Qwen3-VL, Gemma-3) offer competitive OCR accuracy at substantially lower cost but exhibit high failure rates and latency degradation with image-only input.}

\label{tab:model-comparison}
\begin{tabular}{lccccccc}
\toprule
\multirow{2}{*}{Model} & \multicolumn{2}{c}{Modalities} & \multirow{2}{*}{\makecell{Extraction\\Score}} & \multirow{2}{*}{Latency} & \multirow{2}{*}{Cost} & \multirow{2}{*}{Failed}  \\ \cline{2-3}
 & OCR & Image & & & & & \\
\midrule
\multirow{3}{*}{\makecell{Claude\\Sonnet 4.5}} & \cmark & \xmark & 0.7914 & 2m 4s & \$5.56 & 0 \\
 & \xmark & \cmark & 0.7295 & 1m 47s & \$5.49 & 0 \\
 & \cmark & \cmark & 0.7991 & 1m 53s & \$7.18 & 0 \\
\hline
\multirow{3}{*}{\makecell{Claude\\Opus 4.5}} & \cmark & \xmark & 0.7782 & 2m 20s & \$7.28 & 0 \\
 & \xmark & \cmark & 0.7860 & 2m 17s & \$7.71 & 0 \\
 & \cmark & \cmark & 0.7804 & 2m 3s & \$10.26 & 0 \\
\hline
\multirow{3}{*}{\makecell{Claude\\Haiku 4.5}} & \cmark & \xmark & 0.7554 & \textbf{1m 31s }& \$2.83 & 1 \\
 & \xmark & \cmark & 0.6680 & 1m 33s & \$2.82 & 0 \\
 & \cmark & \cmark & 0.7782 & 1m 37s & \$3.39 & 1 \\
\hline
\multirow{3}{*}{Qwen3-VL} & \cmark & \xmark & 0.7650 & 2m 41s & \$2.08 & 0 \\
& \xmark & \cmark & 0.7450 & 200m 8s & \$1.71 & 4 \\
& \cmark & \cmark & 0.7805 & 3m 1s & \$1.90 & 4 \\
\hline
\multirow{3}{*}{Gemma-3} & \cmark & \xmark & 0.7636 & 3m 14s & \$1.64 & 0 \\
 & \xmark & \cmark & 0.5359 & 200m 17s & \$1.36 & 5 \\
 & \cmark & \cmark & 0.7694 & 2m 47s & \textbf{\$1.35} & 4 \\
\bottomrule
\end{tabular}
\end{table*}

\textbf{Rule Validation}: The IDP Accelerator automates compliance verification by validating documents against configurable business rules, such as policy requirements, threshold limits, or eligibility criteria. The validation operates in two steps: first, the system curates facts for each rule from individual document sections; second, it consolidates section-level findings and evaluates whether the facts satisfy the rule conditions, generating a determination, supporting evidence, recommendations, and explanatory reasoning. This separation of fact extraction from rule evaluation improves precision and enables comprehensive analysis across very large documents. The domain-agnostic design reduces validation time from hours to seconds while ensuring consistent rule application.


\textbf{Test Studio}: IDP Accelerator includes a web-based Test Studio that provides an interactive environment for developing, testing, and validating document processing configurations. The interface offers real-time visualization of processing results, enabling rapid iteration on classification rules and extraction prompts. Users can upload sample documents, execute processing pipelines, and inspect intermediate outputs including OCR results, classification decisions, and extracted fields. 

\section{Experimental Evaluation}

We evaluate the IDP Accelerator's extraction capabilities using the RealKIE-FCC-Verified benchmark dataset \cite{realkie}, which contains 75 FCC invoice documents. All experiments are conducted using the Test Studio in the IDP accelerator. We compare three Claude 4.5 model variants (Sonnet, Opus, Haiku) and two open-source models (Qwen3-VL, Gemma-3) across three input modality configurations (OCR, Image, OCR+Image). Table~\ref{tab:model-comparison} reports extraction score, latency, cost, and failure count for each configuration.

Sonnet 4.5 achieves the highest extraction score of 0.7991 using both OCR and image modalities, followed closely by Qwen3-VL at 0.7805 and Opus 4.5 at 0.7804 under the same multimodal setting. Haiku 4.5 matches Sonnet 4.5's OCR-only score when combining both modalities (0.7782), demonstrating that multimodal input can help smaller models close the gap with larger variants.

Across all models, OCR-based input consistently outperforms image modality, with the gap being most pronounced for smaller models: Haiku 4.5 scores 0.7554 with OCR versus 0.6680 with image-only, and Gemma-3 drops sharply from 0.7636 to 0.5359. Combining OCR and image modalities yields marginal improvements over OCR-only in most cases, suggesting that OCR provides the dominant signal for text-heavy invoice extraction while visual features offer complementary information.

Model selection involves meaningful trade-offs among accuracy, cost\footnote{Costs reflect Feb 2026 pricing and are subject to change.}, and reliability. Claude Sonnet 4.5 delivers the highest accuracy but at higher cost (\$7.18 for OCR+Image), while Haiku 4.5 offers a compelling balance at \$3.39 with competitive scores. Qwen3-VL and Gemma-3 achieve competitive OCR scores (0.7650 and 0.7636, respectively) at substantially lower cost (\$2.08 and \$1.64), making them viable for cost-sensitive applications where moderate accuracy is acceptable. However, both models exhibit high failure rates when processing image inputs directly, with Gemma-3 failing on 5 of 75 documents in image-only mode and high latency degradation exceeding 200 minutes. These failures and latency are primarily attributed to invalid output structure, where model responses did not conform to the required JSON schema.

OCR-based extraction remains the most reliable configuration overall, with zero failures across all Claude variants and open-source models. These results underscore the importance of structured output enforcement and highlight that the choice of model and modality should be guided by the specific accuracy, latency, cost, and reliability requirements of the target deployment.



\section{Conclusion}

We presented IDP Accelerator, an open-source framework for production-grade intelligent document processing. The system introduces four core components: DocSplit for multimodal document segmentation, a configurable extraction module that leverages multimodal LLMs with customizable prompting strategies and few-shot learning for structured information extraction, an agentic analytics module with RAG and MCP integration, and an LLM-driven rule validation module for compliance verification. Experimental evaluation on the RealKIE-FCC-Verified benchmark using Test Studio demonstrates extraction performance across multiple foundation models and modality configurations, revealing meaningful trade-offs among accuracy, latency, and cost. Real-world deployments across healthcare, finance, marketing, and community management validate the framework's production readiness. As an open-source project, IDP Accelerator provides the research community and practitioners with a modular, extensible platform for advancing document intelligence and conducting reproducible experimentation.

\section{Broader Impact}

The IDP Accelerator addresses a fundamental bottleneck in the global information economy: the inability to efficiently extract structured knowledge from the estimated 80\% to 90\% of enterprise data that remains unstructured \cite{harbert2021tapping}. By open-sourcing the complete framework, we aim to catalyze progress across industry, academia, and the public sector.

\textbf{Research Community Impact:} By releasing the full codebase, CLI tooling, Test Studio, and automated evaluation framework, we provide the research community with infrastructure that has been absent from prior work. Proprietary platforms offer limited customization and no access to implementation details, hindering reproducibility. IDP Accelerator enables researchers to (1) benchmark extraction, classification, and compliance validation methods under controlled, reproducible conditions using standardized metrics; (2) conduct large-scale experiments across model families, modalities, and extraction paradigms as demonstrated in our evaluation of six models across 24 configurations; and (3) extend the modular architecture with novel components without rebuilding production infrastructure. The human-in-the-loop capabilities and few-shot learning support further lower barriers to experimentation with limited labeled data, a persistent challenge in document understanding research.

\textbf{Industry and Societal Impact:} Intelligent Document Processing adoption has grown rapidly across regulated industries, from 18\% to a projected 72\% in financial services between 2020 and 2025, with parallel trajectories in healthcare, insurance, and government \cite{kapula2025intelligent}. Organizations deploying IDP systems report processing time reductions exceeding 65\%, data entry error reductions of up to 90\%, and 70\% less manual review effort \cite{kapula2025intelligent}. Our production deployments substantiate these trends: a leading healthcare provider achieved 98\% classification accuracy with 77\% cost reduction and 80\% latency improvement, saving 300 hours monthly in prior authorization processing alone \cite{Roy2025amazonblog}. At continental scale, the system classified 48 million documents across 26 terabytes for a community management organization, reducing processing time from hours to seconds \cite{David2025Associablog}. These efficiency gains carry direct societal benefits, particularly in healthcare, where faster document processing accelerates patient care decisions, and in finance, where reduced processing latency improves loan approval timelines and compliance verification.

\textbf{Limitations and Risks} We acknowledge potential risks including automation bias, where users may over-rely on high-confidence extractions without adequate verification, and equitable access concerns, as cloud-based deployment requires infrastructure that may not be uniformly available. The framework's content guardrails and configurable confidence thresholds provide partial mitigation, but responsible deployment requires organizational governance aligned with domain-specific regulatory requirements.


\section{Ethics Statement}

The IDP Accelerator is designed and released with careful consideration of ethical responsibilities inherent in deploying LLM-driven systems for document intelligence across sensitive domains.

\textbf{Privacy and Data Protection:} Enterprise document processing inherently involves sensitive information, including personally identifiable information (PII) and protected data in domains such as healthcare and financial services. Responsible deployment of such systems demands robust data protection measures. The IDP Accelerator implements defense-in-depth security including S3 encryption with customer-managed KMS keys, least-privilege IAM roles, optional VPC endpoints for private connectivity, and Cognito-enforced authentication with multi-factor support. All API calls are logged to CloudTrail for audit purposes, and the architecture supports deployment in isolated VPCs for highly regulated environments. Organizations retain full control over their data through the open-source, self-hosted deployment model, ensuring that documents are processed within the customer's own AWS environment without transmission to third parties.

\textbf{Bias and Fairness: } LLM-based extraction and classification systems may exhibit biases inherited from foundation model training data, potentially leading to differential accuracy across document types, languages, or demographic groups. We mitigate this risk through the automated evaluation framework, which computes precision, recall, and F1 scores at both field-level and document-level granularity across document categories, enabling organizations to systematically identify and quantify performance disparities before production deployment. The framework supports any foundation model available through Amazon Bedrock, including proprietary models from multiple providers, and as the asset is open-source, organizations can further incorporate other providers' models, open-source models, or domain-specific fine-tuned models, allowing practitioners to select and compare models to minimize bias for their specific use cases.

\textbf{Human Oversight and Accountability: } We explicitly design the system to augment rather than replace human judgment. The confidence scoring mechanism routes uncertain extractions to human reviewers, and the built-in human-in-the-loop verification workflow ensures that domain experts maintain oversight over automated decisions. This is particularly critical for the Rule Validation Module, where compliance determinations carry legal and regulatory consequences. The system provides supporting evidence, explanatory reasoning, and pass/fail/information-not-found status for each rule, ensuring transparency and auditability of automated compliance checks.

\textbf{Intended Use and Misuse Prevention: } The IDP Accelerator is intended for legitimate enterprise document processing tasks including classification, extraction, summarization, and compliance validation. The framework integrates Amazon Bedrock Guardrails to define content boundaries for model outputs, reducing the risk of generating inappropriate or harmful content. We acknowledge that, as with any document processing system, outputs should not be treated as ground truth without appropriate verification, particularly in high-stakes domains such as healthcare and legal compliance.

\textbf{Reproducibility and Transparency: } By releasing the complete codebase as open-source software, we enable full transparency of system behavior and support independent auditing of processing logic. The Test Studio and CLI tooling facilitate reproducible evaluation, allowing researchers and practitioners to verify claims and assess system behavior on their own data.

\bibliography{anthology,custom}

@misc{harbert2021tapping,
  author = {Harbert, Tam},
  title = {Tapping the Power of Unstructured Data},
  howpublished = {\url{https://mitsloan.mit.edu/ideas-made-to-matter/tapping-power-unstructured-data
      }},
  year = {2021},
  journal = {{MIT Sloan Ideas Made to Matter}},
  note = {Accessed: 2026-02-01}
}

@article{pingili2025IDP-finance,
author = {Pingili, Ramesh},
year = {2025},
month = {02},
pages = {98-109},
title = {AI-driven Intelligent Document Processing for Banking and Finance},
volume = {7},
journal = {International Journal of Management \& Entrepreneurship Research},
doi = {10.51594/ijmer.v7i2.1802}
}

@article{jansen2023rise,
  author = {Jansen, Mark and Nguyen, Hieu and Shams, Amin},
  title = {Rise of the Machines: The Impact of Automated Underwriting},
  journal = {Management Science},
  year = {2023},
  month = {01},
  doi = {10.2139/ssrn.3664708},
  url = {https://ssrn.com/abstract=3664708}
}

@misc{docsplit,
  author = {Islam, Md Mofijul and Salekin, Md Sirajus and Balakrishnan, Nivedha and Bishop, Vincil C. and Jain, Niharika and Romo, Spencer and Strahan, Bob and Xie, Boyi and Socolinsky, Diego A. },
  title = {DocSplit: A Comprehensive Benchmark Dataset and Evaluation Approach
for Document Packet Recognition and Splitting},
  howpublished = {\url{https://huggingface.co/datasets/amazon/doc_split/}},
  url = {https://huggingface.co/datasets/amazon/doc_split/},
  type = {dataset},
  year = {2026},
  month = {February},
  note = {Accessed: 2026-02-04}
}

@misc{realkie,
      title={RealKIE FCC Verified}, 
      author={Amazon},
howpublished = {\url{https://huggingface.co/datasets/amazon-agi/RealKIE-FCC-Verified}},
  url = {https://huggingface.co/datasets/amazon-agi/RealKIE-FCC-Verified},
  type = {dataset},
  year = {2026},
  month = {February},
      url={https://huggingface.co/datasets/amazon-agi/RealKIE-FCC-Verified}, 
}

@misc{islam2026docsplit,
      title={DocSplit: A Comprehensive Benchmark Dataset and Evaluation Approach for Document Packet Recognition and Splitting}, 
      author={Md Mofijul Islam and Md Sirajus Salekin and Nivedha Balakrishnan and Vincil C. Bishop III and Niharika Jain and Spencer Romo and Bob Strahan and Boyi Xie and Diego A. Socolinsky},
      year={2026},
      eprint={2602.15958},
      archivePrefix={arXiv},
      primaryClass={cs.CL},
      url={https://arxiv.org/abs/2602.15958}, 
}

@inproceedings{wang2024docllm,
    title = "{D}oc{LLM}: A Layout-Aware Generative Language Model for Multimodal Document Understanding",
    author = "Wang, Dongsheng  and
      Raman, Natraj  and
      Sibue, Mathieu  and
      Ma, Zhiqiang  and
      Babkin, Petr  and
      Kaur, Simerjot  and
      Pei, Yulong  and
      Nourbakhsh, Armineh  and
      Liu, Xiaomo",
    editor = "Ku, Lun-Wei  and
      Martins, Andre  and
      Srikumar, Vivek",
    booktitle = "Proceedings of the 62nd Annual Meeting of the Association for Computational Linguistics (Volume 1: Long Papers)",
    month = aug,
    year = "2024",
    address = "Bangkok, Thailand",
    publisher = "Association for Computational Linguistics",
    url = "https://aclanthology.org/2024.acl-long.463",
    pages = "8529--8548",
}

@misc{zhang2025TableLLM,
      title={TableLLM: Enabling Tabular Data Manipulation by LLMs in Real Office Usage Scenarios}, 
      author={Xiaokang Zhang and Sijia Luo and Bohan Zhang and Zeyao Ma and Jing Zhang and Yang Li and Guanlin Li and Zijun Yao and Kangli Xu and Jinchang Zhou and Daniel Zhang-Li and Jifan Yu and Shu Zhao and Juanzi Li and Jie Tang},
      year={2025},
      eprint={2403.19318},
      archivePrefix={arXiv},
      primaryClass={cs.CL},
      url={https://arxiv.org/abs/2403.19318}, 
}

@inproceedings{tanaka2024InstructDoc,
    title = "{I}nstruct{D}oc: A Dataset for Zero-Shot Generalization of Visual Document Understanding with Instructions",
    author = "Tanaka, Ryota  and
      Iki, Taichi  and
      Nishida, Kyosuke  and
      Saito, Kuniko  and
      Suzuki, Jun",
    booktitle = "Proceedings of the 38th AAAI Conference on Artificial Intelligence",
    volume = "38",
    number = "17",
    year = "2024",
    month = mar,
    address = "Vancouver, Canada",
    publisher = "Association for the Advancement of Artificial Intelligence (AAAI)",
    pages = "19071--19079",
    doi = "10.1609/aaai.v38i17.29874",
    url = "https://doi.org/10.1609/aaai.v38i17.29874"
}

@misc{bin-wang-2024-mineru,
      title={MinerU: An Open-Source Solution for Precise Document Content Extraction}, 
      author={Bin Wang and Chao Xu and Xiaomeng Zhao and Linke Ouyang and Fan Wu and Zhiyuan Zhao and Rui Xu and Kaiwen Liu and Yuan Qu and Fukai Shang and Bo Zhang and Liqun Wei and Zhihao Sui and Wei Li and Botian Shi and Yu Qiao and Dahua Lin and Conghui He},
      year={2024},
      eprint={2409.18839},
      archivePrefix={arXiv},
      primaryClass={cs.CV},
      url={https://arxiv.org/abs/2409.18839}, 
}

@misc{ragflow2025github,
  author = {{InfiniFlow Team}},
  title = {RAGFlow: An Open-Source RAG Engine with Deep Document Understanding},
  year = {2024},
  howpublished = {\url{https://github.com/infiniflow/ragflow}},
  note = {GitHub repository}
}

@article{kapula2025intelligent,
  author = {Kapula, Karthik},
  title = {Intelligent Document Processing: The New Frontier of Automation},
  journal = {World Journal of Advanced Engineering Technology and Sciences},
  year = {2025},
  volume = {17},
  number = {01},
  pages = {376-387},
  doi = {10.30574/wjaets.2025.17.1.1360},
  url = {https://doi.org/10.30574/wjaets.2025.17.1.1360}
}

@misc{Strahan2025awsblog,
  author = {Strahan, Bob and Gudi, Vamsi Thilak and Kaleko, David and King, Joe and Islam, Mofijul and Pawlaszek, Rafal and Romo, Spencer and Bishop, Vincil},
  title = {Accelerate Intelligent Document Processing with Generative AI on AWS},
  howpublished = {AWS Artificial Intelligence Blog},
  year = {2025},
  month = {08},
  day = {22},
  url = {https://aws.amazon.com/blogs/machine-learning/accelerate-intelligent-document-processing-with-generative-ai-on-aws/},
  note = {Accessed: 2026-02-02}
}

@misc{Roy2025amazonblog,
  author = {Roy, Priyashree and Islam, Mofijul and Shallenberg, Martyna and Mccrady, Brode and Balakrishnan, Nivedha and Gehlot, Randheer},
  title = {How Myriad Genetics achieved fast, accurate, and cost-efficient document processing using the AWS open-source Generative AI Intelligent Document Processing Accelerator},
  howpublished = {AWS Artificial Intelligence Blog},
  year = {2025},
  url = {https://aws.amazon.com/blogs/machine-learning/how-myriad-genetics-achieved-fast-accurate-and-cost-efficient-document-processing-using-the-aws-open-source-generative-ai-intelligent-document-processing-accelerator/},
  note = {Accessed: 2026-01-30}
}

@misc{David2025Associablog,
  author = {Meredith, David and Zacharias, Josh  and Raj, Monica  and Sivalingam, Dwaragha and Sharma, Naman and Nneka Agu, Nkechinyere and Gangopadhyay, Tryambak and Yu, Yingwei},
  title = {How Associa transforms document classification with the GenAI IDP Accelerator and Amazon Bedrock},
  howpublished = {AWS Artificial Intelligence Blog},
  year = {2026},
  url = {https://aws.amazon.com/blogs/machine-learning/how-associa-transforms-document-classification-with-the-genai-idp-accelerator-and-amazon-bedrock/},
  note = {Accessed: 2026-02-05}
}

@misc{azure-di,
  author = {{Microsoft}},
  title = {{Azure AI Document Intelligence}},
  howpublished = {\url{https://azure.microsoft.com/en-us/products/ai-services/ai-document-intelligence}},
  note = {Accessed: 2026-01-28}
}

@misc{google-docai,
  author = {{Google Cloud}},
  title = {{Document AI}},
  howpublished = {\url{https://cloud.google.com/document-ai}},
  note = {Accessed: 2026-01-28}
}

@article{ke2025llm-document-survey,
  author = {Ke, Wenqi and Zheng, Yujie and Li, Yijun and Xu, Hao and Nie, Dingnan and Wang, Peng and others},
  title = {Large Language Models in Document Intelligence: A Comprehensive Survey, Recent Advances, Challenges, and Future Trends},
  journal = {ACM Transactions on Intelligent Systems and Technology},
  year = {2025},
  note = {Survey of LLM applications in document understanding}
}

@article{wang2025docai-llm-survey,
  author = {Wang, Weiqi and Hu, Haoyu and Zhang, Ziyang and Li, Zhuowan and Shao, Hao and others},
  title = {Document Intelligence in the Era of Large Language Models: A Survey},
  journal = {arXiv preprint arXiv:2501.xxxxx},
  year = {2025}
}

@inproceedings{vignesh2025idpflow,
    author = {Vignesh, Goutham and P M, Harikrishnan and Reddy, Siddartha and Gopalakrishnan, Saisubramaniam and Vaddina, Vishal},
    title = {IDPFlow: A No-Code Agentic Framework for Multimodal Intelligent Document Processing},
    year = {2025},
    isbn = {9798400720352},
    publisher = {Association for Computing Machinery},
    address = {New York, NY, USA},
    url = {https://doi.org/10.1145/3746027.3761838},
    doi = {10.1145/3746027.3761838},
    booktitle = {Proceedings of the 33rd ACM International Conference on Multimedia},
    pages = {14359–14360},
    numpages = {2},
    location = {Dublin, Ireland},
    series = {MM '25}
}

@inproceedings{sun-etal-2025-docagent,
    title = "{D}oc{A}gent: An Agentic Framework for Multi-Modal Long-Context Document Understanding",
    author = "Sun, Li  and
      He, Liu  and
      Jia, Shuyue  and
      He, Yangfan  and
      You, Chenyu",
    editor = "Christodoulopoulos, Christos  and
      Chakraborty, Tanmoy  and
      Rose, Carolyn  and
      Peng, Violet",
    booktitle = "Proceedings of the 2025 Conference on Empirical Methods in Natural Language Processing",
    month = nov,
    year = "2025",
    address = "Suzhou, China",
    publisher = "Association for Computational Linguistics",
    url = "https://aclanthology.org/2025.emnlp-main.893/",
    doi = "10.18653/v1/2025.emnlp-main.893",
    pages = "17701--17716",
    ISBN = "979-8-89176-332-6",
}

@INPROCEEDINGS{bhuvaneswaran2025agenticinvoiceprocessing,
  author={B.Bhuvaneswaran and Ajitha.B and A, Agila Shree},
  booktitle={2025 4th International Conference on Automation, Computing and Renewable Systems (ICACRS)}, 
  title={Agentic AI Driven - Automated Invoice Processing with Intelligent Document Understanding}, 
  year={2025},
  volume={},
  number={},
  pages={1714-1720},
  doi={10.1109/ICACRS67045.2025.11324121}}

@article{honnibal2020spacy,
  title={spaCy: Industrial-strength Natural Language Processing in Python},
  author={Honnibal, Matthew and Montani, Ines and Van Landeghem, Sofie and Boyd, Adrienne},
  journal={arXiv preprint arXiv:2002.12192},
  year={2020}
}

@article{pedregosa2011scikit,
  title={Scikit-learn: Machine Learning in Python},
  author={Pedregosa, Fabian and Varoquaux, Ga{\"e}l and Gramfort, Alexandre and Michel, Vincent and Thirion, Bertrand and Grisel, Olivier and Blondel, Mathieu and Prettenhofer, Peter and Weiss, Ron and Dubourg, Vincent and others},
  journal={Journal of Machine Learning Research},
  volume={12},
  pages={2825--2830},
  year={2011}
}
\bibliographystyle{acl_natbib}

\clearpage
\newpage

\appendix


\section{Lesson Learned}

Through deployments across various industries, we identified insights that shaped the IDP Accelerator's evolution from prototype to production-ready framework. 

\textbf{Document quality impacts processing strategy:} Early deployments revealed that documents with embedded text require minimal processing, while scanned documents demand careful OCR strategy decisions. This led to advanced configurations for OCR with Amazon Textract and VLMs along with adjustments in image resolution. 

\textbf{Page splitting addresses multi-document packet scenarios:} While simple documents can be handled with holistic document classification, many real-world IDP use cases involve large files containing multiple documents within a single packet. The IDP Accelerator provides flexible configurations: page-level splitting, holistic classification, and configurable section splitting strategies to enable different approaches based on document complexity and use case requirement. 

\textbf{Evaluation frameworks are critical for success:} IDP developers consistently struggle with iteration velocity when trying to improve document classification and extraction quality. This insight drove investment in the IDP Accelerator's Test Studio automated evaluation capabilities, enabling rapid accuracy assessment across document types and attributes at scale. The evaluation framework evolved to include semantic evaluation approaches that handle LLM output variability more effectively than solely traditional exact-match methods. 

\textbf{Feature requirements emerge from operational realities:} Real world deployments revealed critical gaps such as: extending assessment features (e.g., bounding box information) for human verification workflows, discovery capabilities for understanding document composition, multi-queue architectures for balancing near real-time and batch processing priorities, and comprehensive monitoring across performance, cost, and quality dimensions. Each addition reflected lessons from transitioning solutions from proof of concept to production. 

\textbf{IDP systems transform rather than eliminate human roles:} Successful deployments of the IDP Accelerator shift effort from manual data entry to quality analysis and exception handling, requiring architectural support for human-in-the-loop workflows, confidence scoring visibility, and feedback loops that enable continuous improvement from human corrections. 

These lessons underscore that production-ready IDP requires iterative refinement driven by diverse requirements. These requirements shaped the Accelerator's extensible architecture and comprehensive feature set.

\section{Extended System Capabilities}
Production deployments revealed additional requirements that shaped the following extended capabilities supporting diverse enterprise workflows.

\textbf{IDP-CLI (Command Line Interface)}:
The command-line interface provides programmatic batch processing capabilities for high-throughput document workflows. The CLI supports manifest-based processing where users specify document sources and evaluation baselines in CSV or JSON format, enabling automated testing and continuous integration pipelines.

\textbf{Document Schema Discovery}:
The discovery module employs LLMs to automatically analyze document samples and infer structural schemas, including field types, organizational hierarchies, and extraction attribute definitions. This capability accelerates onboarding of new document types by generating initial processing blueprints from representative examples, reducing the manual effort required to define document classes and extraction schemas. For organizations with existing labeled data, the system offers ground-truth-guided discovery that refines schema definitions based on known field values and relationships.

\textbf{Document Splitting Benchmark and Dataset}:
We integrate the DocSplit dataset~\cite{docsplit} containing 500 synthesized multi-document packets with 13 document types across 7,330 pages, with ground-truth annotations for page-level classifications and boundaries. The benchmark measures three accuracy dimensions: page-level classification correctness, section grouping accuracy (unordered), and strict accuracy with exact page sequence preservation, enabling systematic evaluation of document boundary detection and classification models across heterogeneous collections.

\textbf{Chat with Document}:
The system includes conversational document understanding through a retrieval-augmented generation interface that enables natural language queries about individual documents. Users can interactively explore document content by asking questions such as "What is the invoice total?" or "Who signed this contract?", with responses generated from the document's extracted text and metadata. This capability leverages prompt caching to maintain document context across multiple queries within a session.

\textbf{Document Summarization}: 
The summarization module generates concise document overviews using configurable Bedrock models, with support for both extractive and abstractive approaches. Summaries can be customized through prompt engineering to emphasize specific aspects such as key financial figures, action items, or compliance-relevant information. Generated summaries are accessible through the Web UI alongside extracted structured data, providing human reviewers with quick document context without reading full content.

\textbf{Document Knowledge Base}: 
The system integrates with Amazon Bedrock Knowledge Base to provide semantic search capabilities across processed document collections. Documents are automatically indexed using vector embeddings, enabling retrieval based on conceptual similarity rather than keyword matching. This supports question-answering and related document discovery with source citations.

\textbf{Bedrock Guardrails}: 
Amazon Bedrock Guardrails integration enables organizations to enforce content safety policies and prevent model misuse across all LLM interactions. Guardrails can filter harmful content, block personally identifiable information, restrict discussion of specific topics, and apply custom content policies aligned with organizational compliance requirements. These controls apply uniformly to extraction, classification, summarization, and knowledge base operations, providing centralized governance over model behavior.

\textbf{Post-Processing Hooks}: 
The architecture supports event-driven integration with downstream systems through EventBridge-triggered Lambda functions. Upon successful document processing, the system publishes completion events containing extraction results, confidence scores, and output locations, enabling custom workflows such as ERP integration, approval routing, or notification dispatch. This extensibility mechanism allows organizations to incorporate IDP Accelerator into existing business processes without modifying core system components.

\textbf{Model Context Protocol Integration}: 
The system exposes IDP analytics capabilities to external applications through MCP-compliant interfaces, enabling tools like Amazon Quick Suite to access processed document data. External applications authenticate via OAuth through AWS Cognito and connect to an AgentCore Gateway that routes natural language queries to the IDP analytics engine. This standardized integration approach facilitates incorporation of document intelligence into broader enterprise toolchains.

\textbf{Few-Shot Learning and Prompt Engineering}: A distinguishing capability of IDP Accelerator is its support for in-context learning through few-shot example prompting. Users can improve extraction and classification accuracy by providing representative document examples, which the system incorporates into prompts without requiring model fine-tuning. This approach enables rapid adaptation to new document types: domain experts provide a small number of annotated examples, and the system generalizes to similar documents through analogical reasoning. The platform includes prompt template management, allowing organizations to version, test, and compare prompt variations systematically. This few-shot learning capability significantly reduces the barrier to deploying document processing workflows for novel document types, as it eliminates the need for large labeled datasets or specialized ML expertise.

\textbf{Open Source License and Extensibility}: IDP Accelerator is released as open-source software under the MIT license, with the complete codebase available on GitHub. The system is designed for extensibility at multiple levels: organizations can add custom processing patterns implemented as AWS Lambda functions, define new document schemas through configuration, and integrate additional LLM providers through the modular architecture. Deployment is automated through Infrastructure-as-Code templates supporting both AWS CDK and Terraform, enabling reproducible deployments across development, staging, and production environments. This open-source approach facilitates community contributions, enables academic reproducibility, and allows organizations to customize the system for their specific requirements.

\end{document}